\documentclass[10pt,twocolumn,letterpaper]{article}

\usepackage{cvpr}              % To produce the CAMERA-READY version
\usepackage{amsmath}
\usepackage[accsupp]{axessibility}
\usepackage{mathtools}
\usepackage{booktabs}
\usepackage{multirow}
\usepackage[normalem]{ulem}
\useunder{\uline}{\ul}{}
\usepackage{caption}
\usepackage{bm}
\usepackage[thinc]{esdiff}
\DeclareMathOperator*{\argmin}{arg\,min}
\usepackage{xcolor}
\usepackage{fancyhdr}
\definecolor{cvprblue}{rgb}{0.21,0.49,0.74}
\usepackage[pagebackref,breaklinks,colorlinks,allcolors=cvprblue]{hyperref}

\def\paperID{EVENTVISION-41} % *** Enter the Paper ID here
\def\confName{CVPR}
\def\confYear{2025}

\title{Spatio-Temporal State Space Model For Efficient Event-Based Optical Flow}

\author{Muhammad Ahmed Humais$^{1}$ ~  Xiaoqian Huang$^{1}$ ~ Hussain Sajwani$^{1}$ ~ Sajid javed$^{2}$ ~ Yahya Zweiri$^{1}$ \\
$^{1}$Advanced Research and Innovation Center (ARIC), Khalifa University, Abu Dhabi, UAE \\
$^{2}$Department of Computer Science, Khalifa University of Science and Technology, Abu Dhabi, UAE
}

\makeatletter
\let\@oldmaketitle\@maketitle% Store \@maketitle
\renewcommand{\@maketitle}{\@oldmaketitle
\centering
\setcounter{figure}{0}
\includegraphics[width=\textwidth]{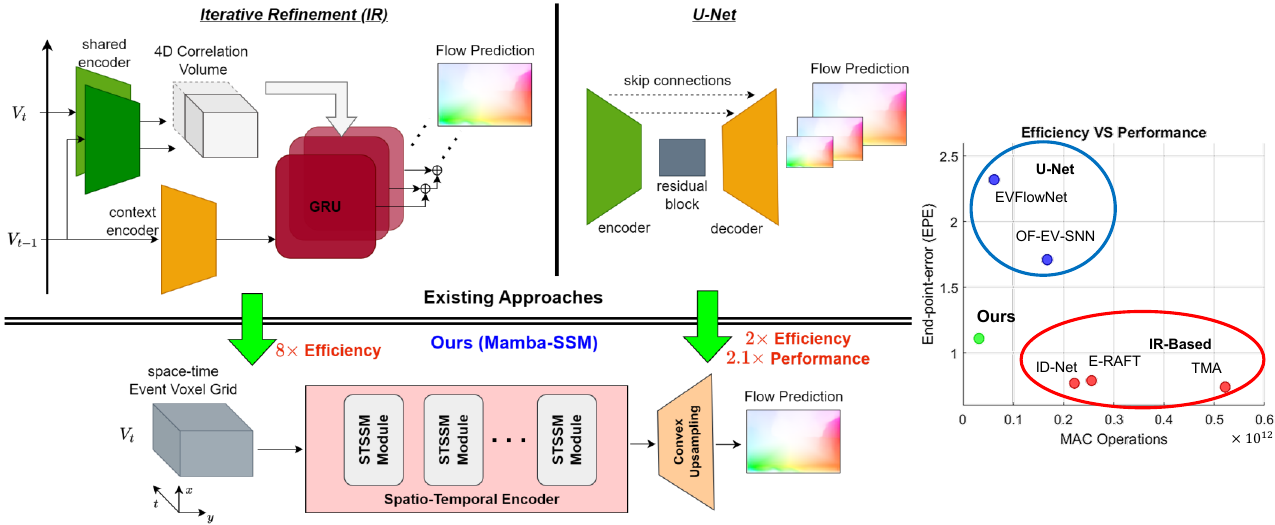}
    \captionof{figure}{\small{\textbf{Top-Left} Previous work used either computationally expensive iterative refinement (IR) framework \cite{eraft, tma} or U-Net architecture with suboptimal performance. \textbf{Bottom-Left} we solved this issue with a novel network architecture (Sec. \ref{sec:network_details}) that leverages our cutting-edge Spatio-Temporal State Space Model (STSSM) module (Sec. \ref{sec:stssm}) for highly efficient spatio-temporal feature extraction and convex upsampling to predict full resolution optical flow with less computations compared to a decoder. \textbf{Right} Comparison of existing approaches in terms of computational efficiency and performance on DSEC public benchmark \cite{gehrig2021dsec}.}}    
    \vspace{0.2cm}
}

\begin{document}

\twocolumn[%
  \begin{@twocolumnfalse}
    \vspace*{-2em}
    \begin{center}
    {\color{gray} 
    This paper has been accepted for publication at the \\ IEEE Conference on Computer Vision and Pattern Recognition (CVPR) Workshops, Nashville, 2025. ©IEEE
    }
    \end{center}
    \vspace{-0.4cm}
    \maketitle
  \end{@twocolumnfalse}
]

\maketitle

\begin{abstract}
Event cameras unlock new frontiers that were previously unthinkable with standard frame-based cameras. One notable example is low-latency motion estimation (optical flow), which is critical for many real-time applications. In such applications, the computational efficiency of algorithms is paramount. Although recent deep learning paradigms such as CNN, RNN, or ViT have shown remarkable performance, they often lack the desired computational efficiency.
Conversely, asynchronous event-based methods including SNNs and GNNs are computationally efficient; however, these approaches fail to capture sufficient spatio-temporal information, a powerful feature required to achieve better performance for optical flow estimation.
In this work, we introduce Spatio-Temporal State Space Model (STSSM) module along with a novel network architecture to develop an extremely efficient solution with competitive performance.
Our STSSM module leverages state-space models to effectively capture spatio-temporal correlations in event data, offering higher performance with lower complexity compared to ViT, CNN-based architectures in similar settings. Our model achieves $4.5\times$ faster inference and $8\times$ lower computations compared to TMA and $2\times$ lower computations compared to EV-FlowNet with competitive performance on the DSEC benchmark. Our code will be available at \texttt{https://github.com/AhmedHumais/E-STMFlow}.

\end{abstract}    
% \begin{figure*}{}
%     \centering
%     \includegraphics[width=1\textwidth]{Figures/intro_fig.pdf}
%     \caption{Comparison With existing approaches}
%     \label{fig:comp}
% \end{figure*}

\section{Introduction}
\label{sec:intro}

Optical flow is a fundamental computer vision problem with numerous practical applications. From frame-based camera perspective, optical flow is estimated between two consecutive frames, hence the maximum update rate is limited by the frame rate of the camera. Due to microsecond temporal resolution, event cameras offer unique advantages, allowing for high update rate and high-bandwidth optical flow estimation \cite{tamingCmax_flow, bflow}.

In contrast to frame-based cameras that encode spatial and color information at a fixed frame rate. Event cameras encode spatio-temporal motion information of edges. Currently, most state-of-the-art approaches obtain frame-like representation of event data \cite{evflownet, spikeflownet, tamingCmax_flow, zhu2019unsupervised, eraft, tma, blinkflow, bflow, sdformerflow} to use CNNs or RNNs which incur huge computational overhead. To address this, sparse and asynchronous computational paradigms have been explored such as Spiking Neural Networks (SNNs) \cite{of_ev_snn, neuro_of_imp} and Graph Neural Networks (GNNs) \cite{hugnet}. Although such methods significantly drop the computational overhead, however, the performance they offer is not comparable to state-of-the-art CNN, RNN \cite{eraft, tma, iterative_deblurring} and ViT-based \cite{blinkflow, sdformerflow} methods. In this work, we focus on developing a light-weight solution for event-based optical flow by leveraging state-space models (SSMs) that are better suited for spatio-temporal event data.

In terms of network architecture, existing methods for event-based optical flow can be broadly categorized into two types. The first type includes methods that draw inspiration from U-Net architecture \cite{evflownet, backtobasics_flow}, and sometimes involve SNNs \cite{spikeflownet, neuro_of_imp, hagenaars_spiking, of_ev_snn}, RNNs \cite{hagenaars_spiking, steflownet, tamingCmax_flow} or Vi-T \cite{sdformerflow, sa_flownet}. The other type includes the methods that employ some kind of iterative refinement (IR) inside the network using a recurrent unit such as GRU \cite{eraft, tma, iterative_deblurring, eem_flow, blinkflow, bflow}. IR can be viewed as an optimization framework embedded into a neural network. While effective, this approach typically incurs computational cost that is several times higher than that of U-Net-like architectures \cite{iterative_deblurring, eraft, evflownet}. We depart from traditional network architectures to introduce a novel design that does not rely on multiple iterations, contextual information, or prior flow estimates. Our network delivers competitive performance compared to IR-based designs and outperforms U-Net-like architectures.

Due to microsecond temporal resolution, event cameras offer unparalleled edge in highly dynamic scenes. However, in such scenarios, high throughput of the sensor data needs efficient and fast processing algorithms. With linear scaling in sequence length, structured state space models (SSMs) \cite{s4, s5, mamba} promise huge potential for such rich spatio-temporal event data. SSMs, originally used to model linear time-invariant (LTI) systems, have proven effective in modeling long-range dependencies (LRDs) in a highly efficient manner compared to existing SOTA methods such as CNN, RNN and Transformers. We argue that SSMs, due to their ability of modeling system dynamics, are inherently suitable for processing such kind of dynamic signal with temporal consistency. Our findings support the above argument, as we show that SSMs boast superior performance with much less computations compared to vanilla transformer within the same network architecture.

In this work, we present a novel \textbf{S}patio-\textbf{T}emporal \textbf{SSM} (STSSM) module to address the unique spatio-temporal nature of input event data. Inspired by RAFT \cite{raft}, we introduced convex upsampling for dense optical flow prediction at full-scale resolution. This allows to significantly reduce the computational cost compared to U-Net architecture, where a decoder with skip connections is used for the same purpose. Finally, we share important insights from our findings about structured SSMs such as S4 \cite{s4}, S4D \cite{s4d} and S5 \cite{s5} that model LTI systems compared to Mamba \cite{mamba}, which models linear time-variant (LTV) system. Our results show competitive performance compared to SOTA methods on public DSEC \cite{gehrig2021dsec} dataset.

Our main contributions are summarized below:
\begin{itemize}
    \item We propose a novel network architecture, different from existing U-Net-based or IR-based networks, our model introduces a spatio-temporal encoder for event volumes and uses efficient convex upsampling for full-scale dense flow prediction.
    \item We developed a novel STSSM module to leverage SSMs for extracting spatial and temporal dynamics of the scene, encoded in event data and demonstrate its effectiveness compared to ViT.
    \item Our experimental results demonstrate a substantial reduction in computational cost while maintaining competitive performance relative to state-of-the-art methods. Additionally, our findings offer valuable insights into the use of SSMs for efficient processing of spatio-temporal data. 
\end{itemize}

% When using numerical integration methods, for efficiency of computations, discrete-time state spaces are employed where SSM is discretized with fixed step size \(\Delta\). 

\section{Related Work}
\label{sec:relatedwork}

Optical flow estimation has been the subject of extensive research for frame-based cameras with numerous notable contributions over the years. However, in this section we will limit our discussion to event-based flow approaches and the recent advancements in SSMs for vision tasks.
\subsection{Event-Based Optical Flow}

Event-based optical flow approaches can be broadly classified into two major paradigms. Optimization-based approaches and learning-based approached. Optimization-based approaches mostly build on \textit{contrast maximization} with the idea of finding spatio-temporal warping of events that generate the sharpest image\cite{unifying_cmax, sofas, Shiba22eccv, tamingCmax_flow}. The same idea is applied in various self-supervised learning-based approaches for training the deep neural networks \cite{evflownet, hagenaars_spiking, zhu2019unsupervised}. 

In terms of network design, the learning-based approaches can be classified into two major paradigms. In fact, these paradigms are borrowed from frame-based optical flow approaches; FlowNet \cite{flownetS} and RAFT \cite{raft}. FlowNet proposed a U-Net architecture to predict optical flow from successive frames, which was adapted by EV-FlowNet \cite{evflownet} for events by obtaining a frame-like representation. Later several works were based on the similar architecture with slight modifications \cite{zhu2019unsupervised, hagenaars_spiking, tamingCmax_flow, spikeflownet, AdaptiveSpikeNet, of_ev_snn}. While effective for small pixel displacement, the U-Net architecture struggles when flow magnitude is high. E-RAFT \cite{eraft}, the event-based version of RAFT \cite{raft} solved this issue by incorporating iterative-refinement (IR) framework. Their approach involves using successive event volumes to build a spatial pyramid of 4D correlation volumes (CVs). However, processing full 4D correlation volume is prohibitively expensive, hence iterative refinement guides the \textit{lookup} to extract local spatio-temporal correlations from full 4D CVs. Several state-of-the-art methods \cite{tma, bflow, blinkflow, dcei_flow, eem_flow} build on similar principle as it somehow embeds optimization into the network and offer huge performance advantage over U-Net. Nevertheless, IR-based approaches are computationally very expensive, hence the need for efficient high-performance flow estimation method remains an open problem.

Another line of research, investigated light-weight spiking neural networks (SNNs) and graph neural network (GNNs) \cite{spikeflownet, AdaptiveSpikeNet, hagenaars_spiking, of_ev_snn, snu_o, hugnet}. Though computationally efficient, they fail to provide comparative performance to SOTA IR-based methods.

\subsection{Vision State Space Models}

Structured state space models (SSMs) have rapidly evolved into promising alternative to existing CNN, RNN and Transformer-based methods \cite{s4, s4d, s5, mamba}. They offer linear scaling of computational complexity with sequence length and state-of-the-art performance in finding long-range dependencies (LRDs). Recently, their application in vision tasks has been studied \cite{vision_mamba, vmamba} as well as for video understanding \cite{videomamba}. From event-based vision perspective, Zubic et al. combined SSMs with attention mechanism for event-based object detection task \cite{ssm_events_scaramuzza}. However, purely SSM-based approaches for event vision holds huge potential \cite{schone2024scalable}.

In our study, for the first time we study the use of SSMs for spatio-temporal data. The space-time coordinates of events are discretized to create spatio-temporal voxel grid, where the discretization error and the polarity is encoded \cite{zhu2019unsupervised}. We reveal that SSMs can efficiently handle such data and provide advantage over vanilla transformers in this specific scenario.

% Existing work used 3D convolutions also but reported negligible performance increase compared to compulational overhead.

\section{Methods}
\label{sec:methodology}

In this section, firstly we present optical flow problem from brightness constancy and spatial consistency constraints to establish the need for spatio-temporal correlations. Then we relate the optical flow problem to event generation model of the event camera and introduce state-space models for optical flow (Sec.\ref{sec:prelim}). The above discussion sets the stage for our novel network design (Sec.\ref{sec:network_details}) and STSSM block (Sec.\ref{sec:stssm}).

\begin{figure*}[htbp]
    \centering
    \includegraphics[width=\textwidth]{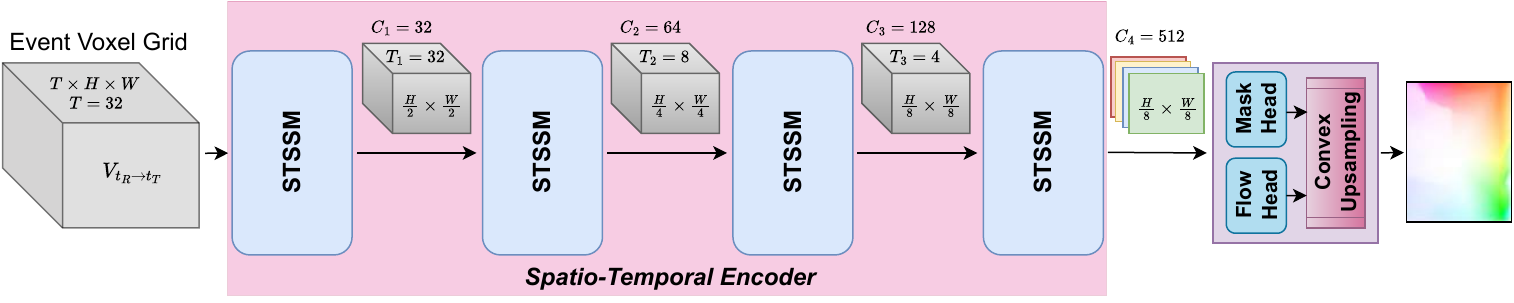}
    \caption{\textbf{Network Architecture} The Spatio-Temporal encoder utilizes successive STSSM Blocks to extract 2D ST feature-maps from input event volume \(V_{t_R \to t_T}\). Then \textit{Flow Head} and \textit{Mask Head}, two-layered CNNs, are used to predict low-resolution flow and mask for convex upsampling to obtain full-scale flow prediction.}
    \label{fig:net_arch}
\end{figure*}

\subsection{Preliminaries}
\label{sec:prelim}
\subsubsection{Optical Flow}
Optical flow estimation relies on the brightness constancy (BC) assumptions. Consider \(\textbf{u} \doteq (u, v)^T\) being pixel coordinate and \(I(\textbf{u}, t)\) being the intensity value at time \(t\), BC is given by:
\begin{equation}
\label{eq:bright_cons}
    I(\textbf{u}, t) = I(\textbf{u} + \Delta \textbf{u}, t + \Delta t),
\end{equation}
where \(\Delta\) denotes small shift in values. First-order approximation of Taylor series expansion of \(I(\textbf{u} + \Delta \textbf{u}, t + \Delta t)\) gives:
\begin{equation}
\label{eq:optical_flow}
    \frac{\partial I}{\partial \textbf{u}} \cdot \textbf{v} = - \frac{\partial I}{\partial t},
\end{equation}
where \(\textbf{v} \doteq \frac{\Delta \textbf{u}}{\Delta t}\) is the optical flow. Above constraint alone is not enough for flow estimation, hence additional constraints are usually imposed, e.g. spatial consistency. For a local event window $\Omega(\textbf{u},t)$, the optical flow can be estimated by solving the optimization problem:
\begin{equation}
\label{eq:spat_consis_of}
    \mathbf{v}^* = \argmin_{\mathbf{v}} \sum_{\textbf{u} \in \Omega} \left\|\nabla L(\textbf{u}) \cdot \mathbf{v} + \frac{\partial L(\textbf{u})}{\partial t}\right\|^2
\end{equation}
This can be solved in closed form using:
\begin{equation}
    \mathbf{v}^* = -(\bm{\Phi}^T \bm{\Phi})^{-1}\bm{\Phi}^T \bm{\Theta},
\end{equation}
where \(\bm{\Phi}\) and \(\bm{\Theta}\) are the matrices of spatial derivatives and temporal derivatives, respectively. This suggests that both spatial and temporal derivatives are essential, prompting the use of 4D correlation volumes in IR-based methods \cite{eraft, tma, bflow, eem_flow}. They also utilize context events, which we argue is redundant in the following discussion.
% Note that the solution would not be possible when the spatial gradient, \(\nabla I = 0\) (no edge) or when the motion is perpendicular to the edge \(\left( \nabla I \: || \: \Delta \textbf{u} \right)\). In such scenarios, additional assumptions such as spatial consistency are used for the flow estimation.

\subsubsection{Event Generation Model}

Event cameras generate asynchronous events. Each event $e_k(t) \doteq (u_k, v_k, t, p_k)$ is a tuple that encodes space-time coordinates \(\textbf{u}_k\), \(t_k\) and the polarity \(p_k \in \{-1, +1\}\) that represents the positive or negative change in the brightness. An event is generated when the change in log intensity \(\Delta L(\textbf{u}_k,t)\) reaches a certain threshold \(\pm C\), i.e.
\begin{equation}
\label{eq:ev_gen1}
    \Delta L(\textbf{u}_k,t) = p_k C.
\end{equation}
Where \(C\) is contrast threshold parameter and \(\Delta t\) is the time elapsed since the last event registered at \(\textbf{u}_k\).  For sufficiently small \(\Delta t\), it has been shown that \( \Delta L \approx \nabla L \cdot \Delta \textbf{u} \) \cite{event_survey}, hence \eqref{eq:ev_gen1} can be rewritten as:
\begin{equation}
\label{eq:ev_gen2}
    \frac{\partial L}{\partial \textbf{u}} \cdot \textbf{v} \Delta t = p_k C
\end{equation}

Equations \eqref{eq:optical_flow} and \eqref{eq:ev_gen2} suggest that events carry motion information. 

% \begin{equation}
%     \nabla L \cdot \mathbf{v} + \frac{\partial L}{\partial t} = 0
% \end{equation}

% This equation can be extended to event cameras by considering the spatiotemporal neighborhood of events. For a local event window $\Omega(x,y,t)$, the optical flow can be estimated by minimizing:

% \begin{equation}
%     \argmin_{\mathbf{v}} \sum_{e \in \Omega} \left\|\nabla L(e) \cdot \mathbf{v} + \frac{\partial L(e)}{\partial t}\right\|^2
% \end{equation}

% where $L(e)$ represents the local intensity gradient estimated from the event stream. This formulation enables direct computation of optical flow from event data, leveraging the camera's high temporal resolution ($\approx 1\mu s$) and avoiding the aperture problem common in traditional frame-based approaches through the integration of multiple events over time.

% Owing to above explanation, we argue that the events in a temporal window for which the optical flow is being estimated should be enough. In contrast, existing approaches\cite (eraft, eflowformer, tma, ofevsnn, sdflowformer) use the events before the reference time for the context, which we argue are redundant for event-based optical flow estimation. In past, the approaches \cite (evflownet) using only the events from \([t_R, t_F]\) temporal window were reported to perform poorly compared to those which utilize context events. However, in this work we show that our novel network design enables the use of only current events for efficient optical flow estimation while providing competitive performance.

\subsubsection{State Space Models}
\label{sec:ssm}
A linear time-invariant (LTI) system can be modeled by a set of first-order differential equations that can be represented in matrix form, called state-space representation. T
% his representation has found to be very useful for modeling and analysis of physical systems because it allows the use of linear algebra. 
A system of \(N^{\text{th}}\) order can be represented by matrices, \(\bm{A} \in \mathbb{R}^{N \times N}\), \(\bm{B} \in \mathbb{R}^{N \times U}\), \(\bm{C} \in \mathbb{R}^{M \times N}\) and \(\bm{D} \in \mathbb{R}^{M \times U}\) as follows:
\begin{equation}
\label{eq:ct-ss}
    \diff{\textbf{x}}{t} = \bm{A}\textbf{x} + \bm{B} \textbf{z} , \quad \quad \textbf{y} = \bm{C} \textbf{x} + \bm{D} \textbf{z}, 
\end{equation}
where \(\textbf{x}(t) \in \mathbb{R}^{N}\) is the state vector, while \(\textbf{z}(t) \in \mathbb{R}^{U}\) and \(\textbf{y}(t) \in \mathbb{R}^{M}\) are the inputs and outputs of the system, respectively. The matrix \(\bm{A}\) is most crucial in determining the dynamics of the system, while the matrices \(\bm{B}\) and \(\bm{C}\) define the input-to-state and state-to-output relations, respectively. Matrix \(\bm{D}\) is often not used as it establishes direct connection from input to output.

The continuous-time model in \eqref{eq:ct-ss} can be \textit{discretized} with a fixed step size \(\Delta\). For example, using zero-order-hold (ZOH) method, the discretized system can be obtained by \(\overline{\bm{A}} = \exp{(\Delta \bm{A})}\) and \(\overline{\bm{B}} = \bm{A}^{-1}(\exp{(\Delta \bm{A})}-\bm{I})\bm{B}\) as follows:
\begin{equation}
\label{eq:d_ssm}
    \textbf{x}_{k+1} = \overline{\bm{A}} \textbf{x}_k + \overline{\bm{B}} \textbf{z}_k , \quad \quad \textbf{y}_k = \bm{C} \textbf{x}_k + \bm{D} \textbf{z}_k.
\end{equation}

Above formulation gives rise to a recurrent model that is also parallelizable. Moreover, usually a structure is imposed on matrix \(\overline{\bm{A}}\) for computational efficiency. Thus, structured SSMs such as S4 \cite{s4} enjoy faster training compared to RNN and significantly better space and time complexity compared to Transformers.

\textbf{Selective State Space Model (Mamba).} Recently, a selection mechanism has been proposed for SSMs that makes the parameters \(\Delta\), \(\bm{\overline{B}}\) and \(\bm{C}\) to be input dependent, making the system to be linear time-variant (LTV). This approach, dubbed as \textit{Mamba} \cite{mamba} is core part of our STSSM module for efficient processing of spatio-temporal event data. We also conduct ablation study to compare with other SSM variants such as S4 \cite{s4}, S4D \cite{s4d} and S5 \cite{s5}. 

As apparent from \eqref{eq:spat_consis_of} that the optical flow estimation requires spatial context along with spatio-temporal correlations. IR-based approaches resort to utilizing two \textit{successive-in-time} views from event data, reference view \(V_R\) and target view \(V_T\) \cite{eraft, blinkflow, bflow}. We argue that given a sufficiently large spatio-temporal window, the events \(\varepsilon(t_R, t_T) \coloneq \langle e(t)|t \in [t_R, t_T] \rangle\) are enough for flow estimation \(F_{t_R \to t_T}\). We propose to use SSMs to extract spatio-temporal correlations along with spatial context only from \(\varepsilon(t_R, t_T)\). We altogether avoid pyramids of 4D correlation volumes that have poor space and time complexity.

\begin{figure*}[htbp]
    \centering
    \includegraphics[width=\textwidth]{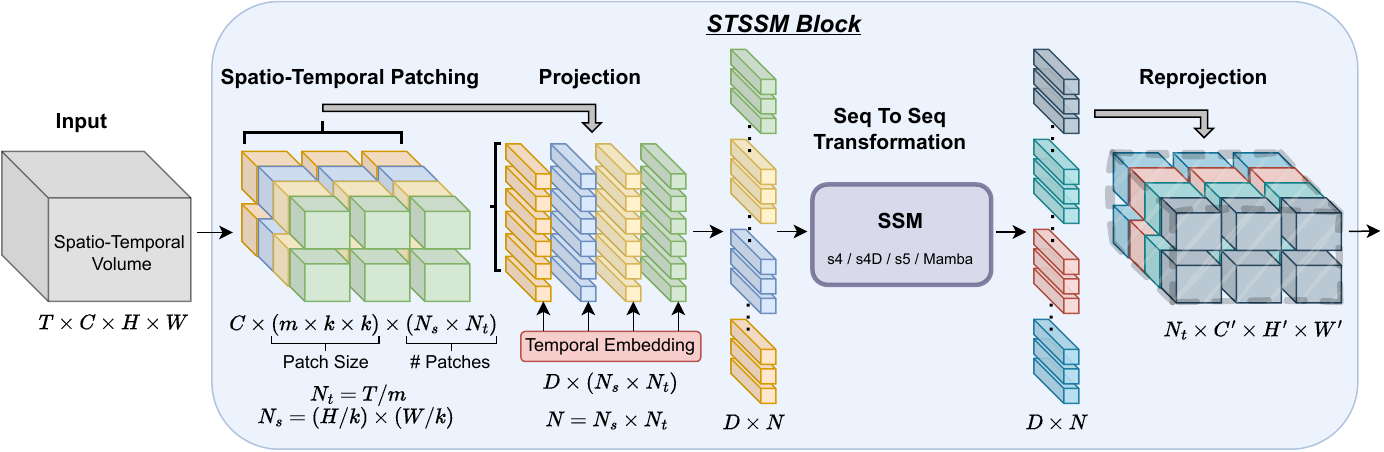}
    \caption{Structure of our STSSM Block. It processes 4D spatio-temporal volumes to extract ST features, keeping the spatio-temporal context intact, while reducing spatial and temporal dimensions to allow rich features with more number of channels.}
    \label{fig:stssm_block}    
\end{figure*}

\subsection{Network Details}
\label{sec:network_details}

First, we describe our network design, which is elemental for developing a highly efficient solution. Our network takes a volume of events as 3D space-time voxel grid. Then our carefully designed spatio-temporal encoder effectively extracts spatio-temporal information through a series of STSSM blocks. Finally, the prediction head takes these features to predict dense flow predictions at full resolution. Fig. \ref{fig:net_arch} shows our network architecture with the necessary details. 

\subsubsection{Input Representation}

We use vocel grid representation, similar to \cite{zhu2019unsupervised, eraft}. For events in a temporal window \(e_i \doteq (u_i, v_i, t_i, p_i) \in \varepsilon (t_R, t_T)\), the voxel grid \(V_{t_R \to t_T} \in \mathbb{R}^{T \times H \times W}\) is encoded as:
\begin{equation}
    V(\Tilde{t},u,v) = \sum_i p_i k_b(u - u_i) k_b(v-v_i) k_b(\Tilde{t}-t^*_i)
\end{equation}
where \(\Tilde{t} \in [0, T-1]\) represents discretized temporal coordinate, and
\begin{equation*}
    t^*_i = (T-1)\frac{t_i - t_R}{t_T-t_R} ,  \quad \quad k_b(a) = \max (0, 1-|a|).
\end{equation*}
In essence, the voxel grid encodes the spatio-temporal discretization error for each space-time coordinate in the event volume, along with the polarity.

\subsubsection{Spatio-Temporal Encoder}

Our spatio-temporal encoder \(\mathcal{F}_{\theta} : \mathbb{R}^{T \times H \times W} \mapsto \mathbb{R}^{C \times \frac{H}{8} \times \frac{W}{8}}\) is consisted of four sequential STSSM blocks. The encoder takes in the spatio-temporal event volume \(V_{t_R \to t_T}\) and encodes the spatio-temporal features into 2D feature maps with \(C\) channels.

Fig. \ref{fig:net_arch} shows the inputs and outputs of each STSSM block. The output hyper volumes are progressively reduced in spatial and temporal dimensions to obtain more compact representations, while the channel dimension is increased to obtain rich feature-set. The patch dimensions in each successive STSSM block is spatially reduced and temporally increased, with \(k=\{32, 8, 4, 1\}\) and \(m=\{1, 4, 2, 4\}\), respectively. This is crucial to obtain fine-grained spatial features with more temporal context. The final STSSM block keeps the spatial dimension of the output to be the same as input, but reduce the temporal dimension to 1, resulting in 2D feature maps.

\subsubsection{Prediction Head}
Traditionally, a decoder network with skip connections is used to upscale the compact encoded features into full-scale predictions. However, they are computationally very heavy. For example, EV-Flownet \cite{evflownet} decoder has \(80 \%\) share in total computational overhead of the network. Instead, we use convex upsampling as given in \cite{raft}. The Flow-head \(\mathcal{H}_F : \mathbb{R}^{C \times \frac{H}{8} \times \frac{W}{8}} \mapsto \mathbb{R}^{2 \times \frac{H}{8} \times \frac{W}{8}}\) provides the flow estimate at \(1/8\) resolution. The Mask-head \(\mathcal{H}_M : \mathbb{R}^{C \times \frac{H}{8} \times \frac{W}{8}} \mapsto \mathbb{R}^{576 \times \frac{H}{8} \times \frac{W}{8}}\) provides \((9 \times 8 \times 8)\) masks to create \(8 \times 8\) patches using weighted average from \(3 \times 3\) neighborhoods of low resolution flow.  

\subsection{STSSM Block}
\label{sec:stssm}
In this section we explain the spatio-temporal SSM (STSSM) block, which is the core novelty of this work. This block is specifically designed to deal with spatio-temporal volumes. Unlike 4D correlation volumes used in SOTA IR-based approaches \cite{raft, eraft, bflow, tma} to obtain spatio-temporal (ST) correlations from 2 views, we use a single ST volume and capture the ST correlations within the ST volume by using SSMs with a carefully designed spatio-temporal patching, projection and reprojection schemes. Fig. \ref{fig:stssm_block} summarizes the working of our STSSM block. 
\subsubsection{Spatio-Temporal Patching}

Unlike video data, where each frame is a snapshot in time, the ST event volume (voxel grid) is sparse for slow motion and dense for fast motion. Hence, a single temporal bin does not necessarily contain all the spatial context. Therefore, to address this we introduce the idea of spatio-temporal patching. For an input of \(\mathcal{I} \in \mathbb{R}^{C \times T \times H \times W}\), we create 3D patches of size \((m \times k \times k)\) for each of \(C\) channels, where \(m\) is the temporal dimension of the patch. 
% In particular, we choose smaller \(m\) with bigger spatial patch size to deal with fast motion. While a larger \(m\) is used to get enough spatial context when there is slow motion. Additionally, \(m\) can also be considered as temporal discretization parameter, where larger \(m\) means coarse discretization and vice versa. 
The temporal dimension of output volume is given by \(N_t = T/m\), with \(T\) being the input temporal dimension.

\subsubsection{Projection}

Similar to ViT, the purpose of projection is to obtain 1D features from the ST patches. This is done via a linear layer \(\mathcal{L}_\theta : \mathbb{R}^{C \times P \times N} \mapsto \mathbb{R}^{D \times N}\), where \(P\) is the patch size and \(N \doteq N_s \times N_t\) is the number of patches, with \(N_s, N_t\) being the number of patches in spatial and temporal dimensions, respectively.

\subsubsection{Temporal Embedding}

Considering SSMs to be LTI systems that map an input sequence to an output sequence through recurrence, as described in \eqref{eq:d_ssm}, the input signal should be in the same domain. However, our input consists of interleaved spatial and temporal domains. We untangle the spatial and temporal domains by introducing temporal embeddings, as illustrated in Fig. \ref{fig:stssm_block}. Specifically, learnable parameters \(\tau_\theta \in \mathbb{R}^{N_t \times D}\) are incorporated along the temporal dimension of the projected ST patches. This approach is analogous to the position encoding used in ViT. Our ablation study shows that position encoding is not needed and introduces unnecessary complication.

\subsection{Seq-to-Seq Transformation}

Recently, Mamba architecture has shown exceptional performance for sequence modeling across various modalities, such as language, audio, video and genomics \cite{mamba}. Mamba boasts linear scaling in sequence length, making them far more efficient than transformers or other methods. Inspired by this breakthrough, we aim to leverage Mamba for capturing spatio-temporal correlations for high efficiency compared to existing methods that rely on 4D correlation volumes. We also conduct a study to compare Mamba with other SSM variants as well as transformers in similar settings in Sec. \ref{sec:ablation}.  

\subsection{Reprojection}

The output sequence from SSMs is reprojected to obtain a spatio-temporal volume from the input sequence data. each vector of size \(D \times 1\) in the input sequence is reprojected to a 2D non-overlapping patch using \(C'\) kernels of size \((D \times l \times l)\), where \(l\) defines the spatial upscaling factor, while the temporal dimension remains the same. Hence, the reprojection module \(\mathcal{R}_\theta : \mathbb{R}^{D \times (N_s \cdot N_t)} \mapsto \mathbb{R}^{N_t \times C' \times H' \times W'}\) transforms the sequence back to spatio-temporal volume. Note that \(H' = (H \cdot l)/k\) and \(W' = (W \cdot l)/k\), therefore, spatial scaling of the output is determined by the parameters \(l\) and \(k\).

\section{Experiments}
% We evaluate our method on DSEC \cite{gehrig2021dsec} optical flow benchmark and provide a detailed comparison with SOTA method in terms of performance, computational efficiency and inference speed (\ref{sec:benchmark}). Furthermore, we perform ablation studies to validate the effectiveness of our designed network architecture, specifically STSSM module.

\subsection{Implementation Details}
\label{sec:impl_det}
We used Mamba \cite{mamba} PyTorch implementation for all of our STSSM modules in the network. In particular, 2 Mamba blocks were used in series with residual connections. The effectiveness of having multiple blocks in series is studied in ablations (Sec. \ref{sec:ablation}).

\textbf{Dataset.} We used DSEC optical flow dataset \cite{gehrig2021dsec} in this work. It comprises of driving sequences in both day and night times. Ground truth for test data is not available, so we created a validation split following \cite{of_ev_snn} for training and ablation studies. we apply random vertical and horizontal flipping and random time-scaling for data augmentations.

\textbf{Training Details.} Our network is implemented using Pytorch, trained using RTX 4090 GPUs with a batch size of 10. AdamW is used as the optimizer with a learning rate of $2\times 10^{-4}$ and a decay rate of $1\times 10^{-5}$ for 100 epochs. Additionally, the OneCycle LR scheduler is used for better convergence. Similar to previous works, we used L1 loss for all our networks. 
\begin{equation}
\label{eq:loss1}
    \mathcal{L} =\frac{1}{|\mathcal{V}|}\sum_{(x,y) \in \mathcal{V}} {||F_j^{gt}(x,y) - F_j^{pred} (x,y)||}_1
\end{equation}
where, \(\mathcal{V}\) is the set of pixels with a valid ground truth flow. 
% This is needed as the ground truth is not available for all pixels in the frame. 

\begin{table}[htbp]
\resizebox{\columnwidth}{!}{%
\begin{tabular}{lcccccc}
\hline
\multicolumn{1}{c}{\multirow{2}{*}{}}                & \multirow{2}{*}{Type} & \multicolumn{4}{c}{Performance}                              & \multirow{2}{*}{GMAC} \\ \cline{3-6}
\multicolumn{1}{c}{}                                 &                       & EPE           & AE            & 1PE           & 3PE          &                       \\ \hline
MultiCM \cite{Shiba22eccv}          & Optim.                & 3.47          & 13.98         & 76.6          & 30.9         & -                     \\ \hline
TamingCM \cite{tamingCmax_flow}    & U-Net                 & 2.33          & 10.56         & 68.3          & 17.8         & -                     \\
OF\_EV\_SNN \cite{of_ev_snn}      & U-Net                 & 1.71          & 6.338         & 53.7          & 10.3         & 168                   \\
EV-FlowNet \cite{evflownet, eraft}  & U-Net                 & 2.32          & 7.90          & 55.4          & 18.6         & {\ul 62}              \\ \hline
E-RAFT \cite{eraft}                 & IR                    & 0.79          & {\ul 2.85}    & 12.7          & {\ul 2.7}    & 256                   \\
E-RAFT w/o WS \cite{eraft}          & IR                    & 0.82          & -             & 13.5          & 3.0          & -                     \\
Bezier \cite{bflow}                 & IR                    & {\ul 0.75}    & \textbf{2.68} & 11.9          & 2.4          & -                     \\
TMA \cite{tma}                      & IR                    & \textbf{0.74} & \textbf{2.68} & \textbf{10.9} & {\ul 2.7}    & 522                   \\
ID-Net \cite{iterative_deblurring} & IR                    & 0.77          & 3.00          & {\ul 12.1}    & \textbf{2.0} & 222                   \\ \hline
\textbf{Ours}                                        & SSM                   & 1.11          & 4.29          & 26.4          & 5.0          & \textbf{32}           \\ \hline
\end{tabular}%
}
\caption{Evaluation results on DSEC \cite{gehrig2021dsec} dataset.}
\label{tab:dsec_benchmark}
\end{table}
\vspace{-0.4cm}

\begin{table}[htbp]
% \resizebox{\columnwidth}{!}{%
% \begin{tabular}{lccccccc}
% \hline
%                        & \textit{iters} & EPE  & AE    & 1PE   & 3PE   & GMAC  & Memory  \\ \hline
% \multirow{2}{*}{ERAFT \cite{eraft}} & 1     & 1.71 & 4.74  & 33.79 & 10.87 & 92.2  & {\ul 1.61 GB} \\
%                        & 2     & 1.18 & {\ul 3.68}  & {\ul 21.24} & 5.98  & 107.1 & 1.74 GB \\ \hline
% \multirow{2}{*}{TMA \cite{tma}}   & 1     & 2.65 & 10.43 & 33.39 & 17.09 & 158.2 & 3.40 GB \\
%                        & 2     & \textbf{0.81} & \textbf{2.88}  & \textbf{12.84} & \textbf{2.87}  & 233.6 & 4.32 GB \\ \hline
% \textbf{Ours}          & 1     & {\ul 1.11} & 4.23  & 26.41 & {\ul 5.00}  & \textbf{32.4}  & \textbf{1.2 GB}  \\ \hline
% \end{tabular}%
% }
\centering
\resizebox{\columnwidth}{!}{%
\begin{tabular}{lcccccc}
\hline
                       & \textit{iters} & EPE           & AE            & 3PE           & GMAC          & Memory          \\ \hline
\multirow{2}{*}{ERAFT \cite{eraft}} & 1              & 1.71          & 4.74          & 10.87         & 92.2          & {\ul 1.61 GB}   \\
                       & 2              & 1.18          & {\ul 3.68}    & 5.98          & 107.1         & 1.74 GB         \\ \hline
\multirow{2}{*}{TMA \cite{tma}}   & 1              & 2.65          & 10.43         & 17.09         & 158.2         & 3.40 GB         \\
                       & 2              & \textbf{0.81} & \textbf{2.88} & \textbf{2.87} & 233.6         & 4.32 GB         \\ \hline
\textbf{Ours}          & 1              & {\ul 1.11}    & 4.23          & {\ul 5.00}    & \textbf{32.4} & \textbf{1.2 GB} \\ \hline
\end{tabular}%
}
\caption{Comparison with high-performance iterative methods with reduced number of iterations for fair comparison with comparable GMACs.}
\label{tab:it_comp}
\end{table}
\vspace{-0.5cm}
\begin{figure*}[ht]
    \centering
    \includegraphics[width=0.97\textwidth]{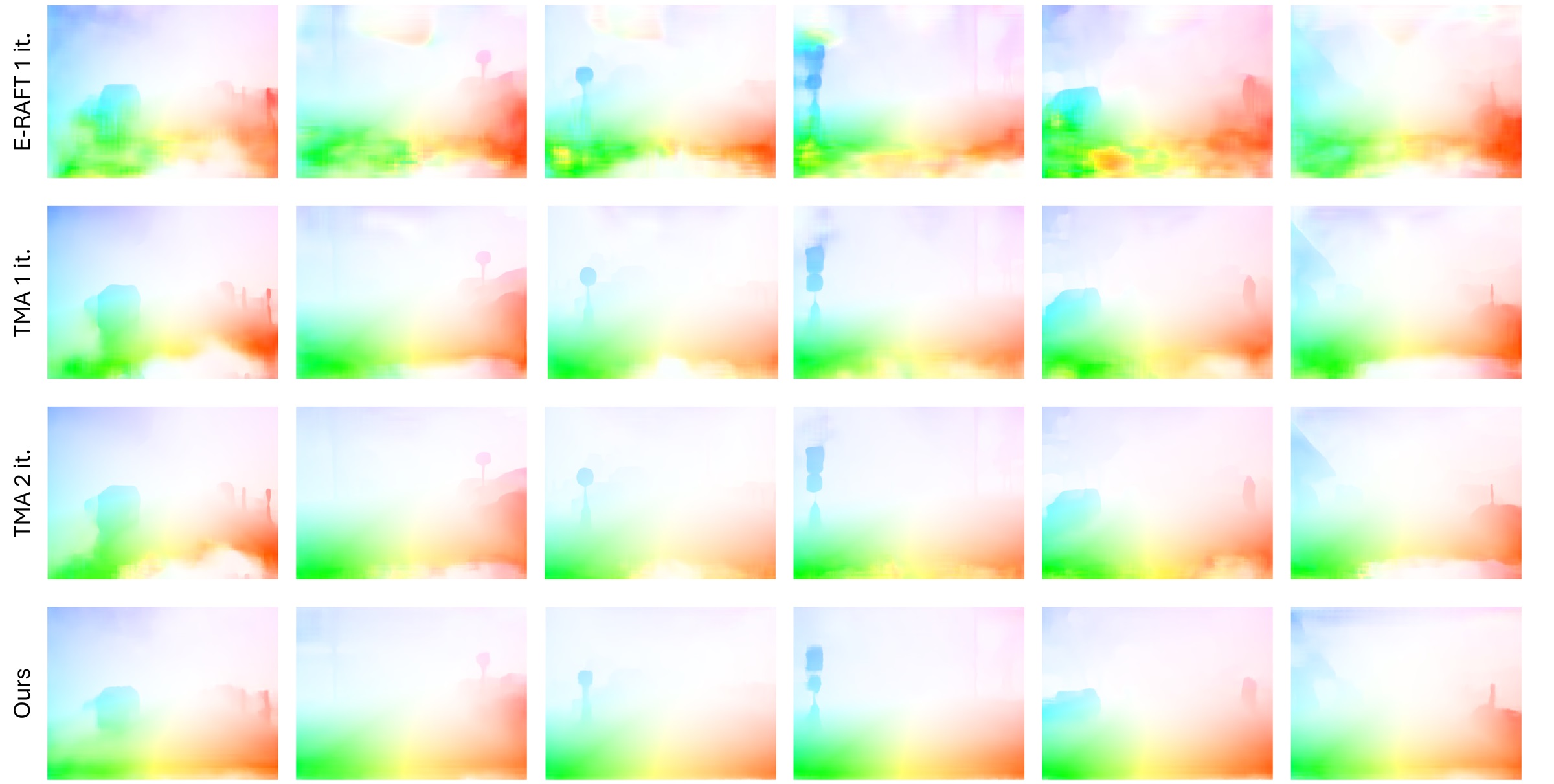}
    \caption{Optical flow prediction of our model on the DSEC dataset compared ground truth compared to state-of-the-art ERAFT \cite{eraft} and TMA \cite{tma} methods. For fair comparison, their 1 iteration versions were used to generate the results. As ground truth is not available for test set, we also show the results using TMA (2 iters) for reference as it is closer to ground truth.}
    \label{fig:dsec_qual}
\end{figure*}

\begin{figure*}[htbp]
    \centering
    \includegraphics[width=\textwidth]{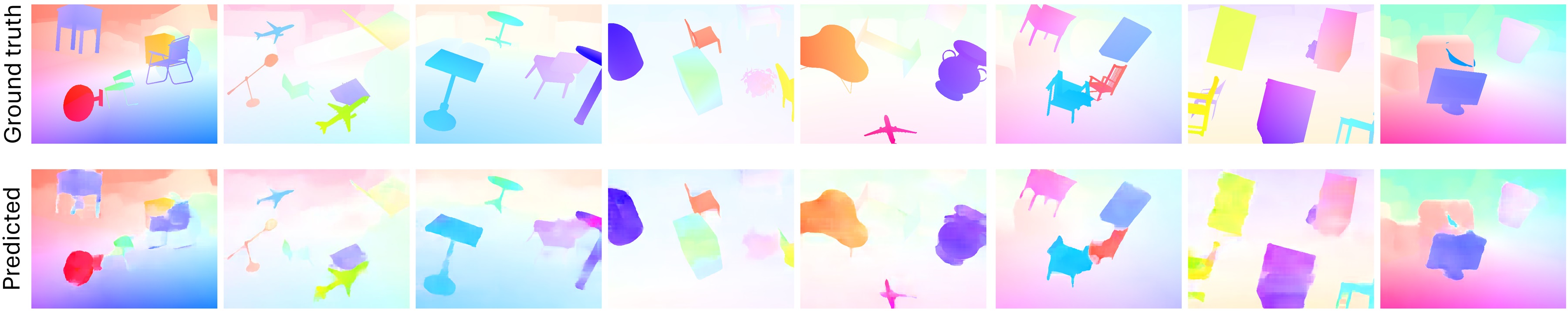}
    \caption{Optical flow prediction of our model on the BlinkFlow dataset compared ground truth. We demostrate our model's performance on cases with multiple objects, various directions, and occlusions.}
    \label{fig:qual}
\vspace{-0.4cm}
\end{figure*}

\subsection{Benchmark Comparisons}
\label{sec:benchmark}
In this section, we provide a comparison with SOTA approaches for event-based optical flow. Table \ref{tab:dsec_benchmark} summarizes the comparison results. We divide the existing SOTA methods according to the type of network they employ. Optimization-based approaches usually perform inferior to learning-based approaches (also can be checked from the DSEC website). This can be explained due to the explicit assumptions on which the optimization frameworks rely. 
% However, they are advantageous because of their independency from the training data. 
% Secondly, self-supervised methods also suffer from similar issue due to the supervisory signal (Loss function) stemming from certain assumptions which may not necessarily be valid for such practical scenarios. 

% The results show the dominance of \textit{iterative refinement} (IR)-based approaches on U-Net-based approaches. This in part could be due to the use of explicit calculation of 4D correlation volumes for spatio-temporal correlations, while U-Net-based approaches implicitly attempt to do the same via CNN layers. Moreover, IR-based approaches are better at handling edge cases where pixels move out of the frame. In addition, iterative mechanism substantially helps in predicting sharper motion boundaries. Our STSSM-based method employs powerful Mamba architecture for efficiently extracting spatio-temporal correlations and uses convex upsampling for efficient full-scale prediction with sharp motion boundaries.

Our method achieves an EPE of 1.11 pixels with a significantly lower computation cost (32 GMACs) compared to others. Compared to supervised EV-FlowNet \cite{eraft, evflownet}, our method offers \(264 \%\) reduction in EPE with \(2\times\) increase in computational efficiency. Compared to state-of-the-art E-RAFT \cite{eraft} and TMA \cite{tma}, our method offers \(8\times\) and \(16\times\) reductions in computational cost with only 0.32 px and 0.37 px increase in error (EPE), respectively.

Since IR-based methods \cite{eraft, tma, iterative_deblurring} perform significantly better (See Table \ref{tab:dsec_benchmark}) at a cost of higher GMACs, we reduced the number of iterations to achieve a fair comparison in Table \ref{tab:it_comp}. Our method achieves significantly lower GMACs and demonstrates a clear performance advantage over state-of-the-art.

\subsection{Qualitative Results}
Fig. \ref{fig:dsec_qual} shows the prediction results on the DSEC test set. Since ground truth is not available, we provide results using TMA \cite{tma} (with 2 iterations) for reference as it features low testing error (see Table \ref{tab:it_comp}). Compared to state-of-the-art ERAFT\cite{eraft} and \cite{tma} (with 1 iteration), our method produces sharper and smoother predictions, even in the presence of high variations in flow magnitude, demonstrating the effectiveness of our STSSM module for spatio-temporal correlations.

Since DSEC does not contain many independently moving object (IMOs) and the motion is fairly coherent, we train and test our method on the synthetic BlinkFlow \cite{blinkflow} dataset. We randomly split the data into \(80\%\) for training and \(20\%\) for testing. Fig. \ref{fig:qual} shows the qualitative results obtained from the testing split, demonstrating the effectiveness of our method in dealing with IMOs; however, we find that it struggles in dealing with occlusion cases and thin objects, but otherwise it provides moderately sharp motion boundaries, even in complex and cluttered scenarios.

\subsection{Ablation Studies}
\label{sec:ablation}
This section provides ablation studies to analyze the effect of our proposed design choices. Since the ground truth for DSEC is not available, we used the split of \cite{of_ev_snn} for ablations, so there may be a discrepancy with the benchmarking results reported in Table \ref{tab:dsec_benchmark}. 
% In this section we present a study to empirically verify our hypothesis for specific design choices. Firstly, we study the effect of incorporating position encoding and temporal encoding within our STSSM block (\ref{sec:enc_comp}). Then we study the performance of different SSM variants when incorporated within our STSSM block (\ref{sec:ssm_var}). Finally, we investigate the effectiveness of our STSSM block for incorporating SSM by comparing with vanilla Transformer baseline (\ref{sec:mamba_vs_vit}). The details of hyper-parameters for all the networks used for ablation are given in supplementary materials.

\begin{table}[]
\centering
\resizebox{0.91\columnwidth}{!}{%
\begin{tabular}{cccclccc}
\hline
\multirow{2}{*}{Enc.} & \multicolumn{3}{c}{Mamba} &  & \multicolumn{3}{c}{Vi-T} \\ \cline{2-4} \cline{6-8} 
                      & EPE    & 3PE   & Params   &  & EPE    & 3PE   & Params  \\ \hline
No                    & 0.90   & 2.7   & 8.68M    &  & 1.36   & 7.6   & 14.1M    \\
t                     & 0.88   & 2.6   & 8.68M    &  & 1.34   & 7.5   & 14.1M   \\
t+p                   & 0.89   & 2.6   & 11.53M   &  & 1.15   & 4.8   & 16.9M   \\ \hline
\end{tabular}%
}
\caption{Ablation studies results for incorporating no encoding (No), temporal encoding (t), or both temporal and position encoding (t+p).}
\label{tab:enc_comp}
\end{table}
% \vspace{-0.4cm}

\subsubsection{Temporal Encoding and Position Encoding}
\label{sec:enc_comp}
Position encoding is an elemental part of ViT \cite{vit} and has also been used with visual Mamba architectures \cite{videomamba, vmamba, vision_mamba}. However, we argue in Sec. \ref{sec:stssm} that our STSSM block uses SSM to extract spatio-temporal correlations, hence we only need temporal encoding to separate spatial and temporal domains. The results presented in Table \ref{tab:enc_comp} summarize our findings and supports our hypothesis.
% We validate this hypothesis by training networks with and without temporal encoding as well as with both position and temporal encoding.

% Table \ref{tab:enc_comp} shows the results for ablation. A slight increase in performance is observed when temporal encoding is incorporated. While adding position encoding slightly worsen the performance. In case of no temporal encoding, the network has to learn by itself to discriminate the temporal dimension from the sequence input. However, adding position encoding increases the complexity of the network (2.85 additional parameters), which can make it harder to train. 

% We also study the effect of position and temporal encoding when SSM \cite{mamba} is replaced by ViT \cite{vit} in our STSSM block. Significant improvement in performance is observed when both temporal and position encoding is used with ViT. This is because of the fundamental difference in SSMs and ViTs which is usually overlooked that the SSMs preserves the sequence information like RNNs while Transformers solely depend on encoding to embed the position information.

\begin{table}[htb]
\centering

\resizebox{\columnwidth}{!}{%
\begin{tabular}{lcccccc}
\hline
                       & \(\times N\) & EPE  & 1PE  & 3PE & GMAC & Time\\ \hline
S4 \cite{s4}                    & 1            & 1.1  & 38.1 & 4.2 & \textbf{28.2}                                                 & 20.9 ms                                             \\ \hline
\multirow{2}{*}{S4D \cite{s4d}}   & 1            & 1.07 & 36.4 & 4.0 & \textbf{28.2}                                                & 10.8 ms                                             \\
                       & 2            & 1.3  & 47   & 6.5 & 31.4                                                 & 16.8 ms                                             \\ \hline
\multirow{2}{*}{S5 \cite{s5}}    & 1            & 1.14 & 40   & 4.6 & 29.8                                                 & 7.1 ms                                              \\
                       & 2            & 0.96 & 31   & 3.0 & 34.7                                                 & 12 ms                                               \\ \hline
\multirow{2}{*}{Mamba \cite{mamba}} & 1            & 0.95 & 30   & 3.2 & 28.7                                                 & \textbf{6.3 ms}                                             \\
                       & 2            & \textbf{0.88} & \textbf{26.6} & \textbf{2.6} & 32.4                                                 & 7.7 ms                                              \\ \hline
\end{tabular}%
}
\caption{Ablation studies for different SSM variants with \(N\) blocks in series within each STSSM module.}
\label{tab:ssm_comp}
\end{table}

\vspace{-0.5cm}
\subsubsection{SSM variants}
\label{sec:ssm_var}
% Structured SSMs have evolved from continuous-time model (S4) \cite{s4}, to discrete-time model (S4D) \cite{s4d} and then a more efficient parallel scan version (S5). Finally, with selection mechanism, Mamba achieved an adaptive model, in which the SSM parameters are input-dependent, which helps in dealing with noise effectively \cite{mamba}.  

Table \ref{tab:ssm_comp} shows the results for different SSM variants. \(N \in \{1, 2\}\) blocks in series were used for comparisons. The continuous-time model (S4) \cite{s4} has the worst execution speed, but offers similar performance to the diagonal version (S4D) \cite{s4d}. However, S4D performance drops when multiple blocks are used in series. In contrast, the S5 \cite{s5} and Mamba \cite{mamba} architectures provide better performance when multiple blocks are used in series. We suspect that this is due to the mixing layer present in the S4D architecture, implemented by position-wise linear layer. As expected, Mamba provides the best performance among all SSM variants with high computational efficiency and fast inference speed.

% \usepackage{multirow}

% \begin{table}[htb]
% \begin{tabular}{lcccccc}
% \hline
%                      & N & EPE  & 3PE & Params & GMAC & \begin{tabular}[c]{@{}c@{}}Time\\ (ms)\end{tabular} \\ \hline
% \multirow{2}{*}{Tr.} & 1 & 1.20 & 5.3 & 12.1M  & 50.3 & 18.5                                                \\
%                      & 2 & 1.15 & 4.8 & 16.9M  & 75.7 & 31.8                                                \\ \hline
% \multirow{2}{*}{M}   & 1 & 0.95 & 3.2 & 6.5M   & 28.7 & 6.3                                                 \\
%                      & 2 & 0.88 & 2.6 & 8.7M   & 32.4 & 7.7                                                 \\ \hline
% M+Tr.                & 2 & 0.96 & 3.0 & 14.1M  & 57.0 & 14.2                                                \\ \hline
% \end{tabular}
% \caption{Comparison of Mamba and Transformer for seq-to-seq transformation in our STSSM module. (M+Tr.) refers to the network with the last STSSM block of the network uses Transformer and the rest use Mamba.}
% \label{tab:mamba_vit}
% \end{table}

\begin{table}[]
\resizebox{\columnwidth}{!}{%
\begin{tabular}{lcclccc}
\hline
                       & N & EPE  & 1PE                      & 3PE & Params & GMAC  \\ \hline
Conv (Baseline)        & - & 1.08 & 31.6 & 3.3 & \textbf{3.7M}   & 40.1  \\
Conv 3D                & - & 0.91 & \textbf{26.4} & 3.2 & 9.1M   & 292.8 \\ \hline
\multirow{2}{*}{Vi-T}  & 1 & 1.20 & 42.0                     & 5.3 & 12.1M  & 50.3  \\
                       & 2 & 1.15 & 40.0                     & 4.8 & 16.9M  & 75.7  \\ \hline
\multirow{2}{*}{Mamba} & 1 & 0.95 & 30.0                     & 3.2 & 6.5M   & 28.7  \\
                       & 2 & \textbf{0.88} & 26.6                     & 2.6 & 8.7M   & \textbf{32.4}  \\ \hline
Mamba + Vi-T           & 2 & 0.96 & 30.7                     & 3.0 & 14.1M  & 57.0  \\ \hline
\end{tabular}%
}
\caption{Comparison of Mamba and Transformer for seq-to-seq transformation in our STSSM module. (Mamba + Vi-T) refers to the network with the last STSSM block of the network uses Transformer and the rest use Mamba. \(N\) is the number of Mamba/Vi-T modules per STSSM block.}
\label{tab:mamba_vit}
\end{table}
% \vspace{-0.4cm}
\subsubsection{CNN vs Mamba vs Transformer}
\label{sec:mamba_vs_vit}
Our STSSM module is specifically designed to exploit SSMs to extract spatio-temporal correlations from spatio-temporal input data. In fact, 2D or 3D CNNs can also be used for the same purpose. We obtain results using different backbones (encoders) to specifically study the effectiveness of Mamba in STSSM. Conv (Baseline) refers to the encoder from ERAFT \cite{eraft}, We also developed a 3D-CNN based encoder (referred to as Conv 3D), though more effective compared to the Baseline, it results in an exponential increase in computational cost (See Table \ref{tab:mamba_vit}). This explains how Mamba is extremely efficient for such task, given its linear scaling of complexity. We also replace Mamba with the transformer encoder \cite{transformer} with 2 heads (referred to as Vi-T). However, transformer introduces unnecessary complexity, making it prone to overfitting and harder to train. In fact, the training curves obtained show noticeable noise, validating our hypothesis.
% SSMs, based on the learned dynamic model, transform an input sequence to an output sequence as explained in Sec. \ref{sec:ssm}. We exploit this property to extract ST features with spatial and temporal context embedded in the output. Unlike SSMs, transformers \cite{transformer} utilize attention mechanism and only encode the interdependency of the elements in input sequence. 

% To validate our hypothesis, we conduct a comparative study where we replace Mamba in our STSSM module with vanilla transformer encoder block \cite{transformer}. The results in Table \ref{tab:mamba_vit} show significant performance drop for the transformer-based implementation with much higher parameters and computational complexity. Moreover, the training loss curve exhibit high noise (see supplementary materials), further validating the unsuitability of transformers in this particular setup.

\section{Conclusion}
\label{sec:Conclusion}

In this work, we introduce the powerful Mamba SSM for the first time for event-based optical flow. Our findings suggest that Mamba, and in general SSMs can be a good fit for processing spatio-temporal data such as event data. We developed a novel STSSM module that efficiently and effectively captures ST correlations and can also be cascaded to get compact representation with rich feature-set. We incorporate this module into our novel state-of-the-art network architecture designed for dense optical flow predictions from event volumes (space-time voxel grids). This results in the most efficient network to date that provides performance comparable to that of SOTA methods. Our findings reveal that Mamba is more suitable compared to transformers for extracting spatio-temporal correlations. They perform similarly to 3D CNNs but with exceptionally high computational efficiency for this particular task.
\section*{Acknowledgements}
This work is supported by Sandooq Al Watan under Grant SWARD-S22-015, STRATA Manufacturing PJSC, and
Advanced Research and Innovation Center (ARIC), which is jointly funded by Aerospace Holding Company LLC, a wholly-owned subsidiary of Mubadala Investment Company PJSC and Khalifa University for Science and Technology.
{
    \small
    \bibliographystyle{ieeenat_fullname}
    \bibliography{bibli}
}


\end{document}